\documentclass[sigconf]{acmart}
\AtBeginDocument{%
  }
\renewcommand\footnotetextcopyrightpermission[1]{} 

\setcopyright{none}
\usepackage{booktabs}
\usepackage{multirow}
\usepackage{makecell}
\usepackage{colortbl}
\begin{document}

\title{High-Capacity Robust Watermarking Technology for High-Resolution Images}

\author{Shaowu Wu}
\email{wushw25@mail2.sysu.edu.cn}
\affiliation{%
	\institution{School of Computer Science and
		Engineering, Sun Yat-sen University}
	\city{Guangzhou}
	\state{}
	\country{China}
}
\author{Wei Lu}
\authornote{Corresponding authors.}
\email{luwei3@mail.sysu.edu.cn}
\affiliation{%
	\institution{School of Computer Science and
		Engineering, Sun Yat-sen University}
	\city{Guangzhou}
	\country{China}
}
\author{Qing Qian}
\email{qingqian@mail.gufe.edu.cn}
\affiliation{%
	\institution{School of Information, Guizhou University of Finance and Economics}
	\city{Guiyang}
	\state{}
	\country{China}
}
\author{Mingsen Deng}
\email{msdeng@mail.gufe.edu.cn}
\affiliation{%
	\institution{School of Information, Guizhou University of Finance and Economics}
	\city{Guiyang}
	\state{}
	\country{China}
}



\renewcommand{\shortauthors}{Trovato et al.}

\begin{abstract}
  Most existing watermarking techniques are primarily designed for low-resolution images, with few methods tailored for high-resolution images. Moreover, the embedding capacity is often limited to fixed lengths (e.g., 30, 100, 256 bits, etc.), which struggles to meet practical demands. To address these issues, this paper proposes a high-capacity robust watermarking method for high-resolution images, capable of embedding a watermark of 4 KB (32,768 bits) into images with a resolution of 1024×1024, achieving an embedding rate of 0.0313 bpp. Specifically, this paper adopts a block-wise strategy to effectively embed the watermark into local regions, enabling the network to train and learn normally even under low-resource conditions. The encoder and decoder structures respectively employ a reversible symmetric architecture with three convolutional and three deconvolutional layers, ensuring consistency in the coupling and decoupling of the watermark and image features. Additionally, the loss function combines global and local losses with weighted contributions. By incorporating constraints on the visual quality and robustness of local block regions, the overall imperceptibility and robustness of the image are further enhanced. Extensive experimental results verify that the proposed method is effective and feasible in high-resolution image scenarios with high-capacity watermarking, while demonstrating strong robustness against various noise attacks.

\end{abstract}


\keywords{Digital watermark, copyright protection, robust residual}



\maketitle

\section{Introduction}
With the widespread adoption of large models, AIGC content generation has become an important application scenario for large models \cite{wang2023security,hosny2024digital,Zhang_2024_CVPR,ye2024privacy,chen2025tag,ye2026securing}. Through large models, the work efficiency of various industries has been greatly improved, which has played a role in reducing costs and increasing efficiency for companies, enterprises or users. However, with the convenience of using large models, the copyright ownership and traceability of the generated content have become objects of concern. How to trace and authenticate the AIGC-generated content has become a new challenge.

At present, there is still a lack of appropriate AIGC traceability authentication strategies. Although digital watermarking \cite{nematollahi2017digital,begum2020digital,10123415,zhang2024robust,zhang2024dual,he2024camera,luo2025nerf,rao2025dynmark,wang2024achieving} is a technology that can solve data traceability and copyright authentication, the current digital watermarking technology is difficult to meet the requirements of high-capacity data. For example, the C2PA alliance adds the C2PA manifest (which identifies the traceability information of AIGC services and models) as metadata to the AIGC-generated content \cite{mo2023towards,Hyun_2025_ICCV,heeger2025eu}. The C2PA manifest has gradually become an actual standard, and the embedded metadata is high-capacity watermark information. At the same time, the image carrier is no longer low-resolution images, but high-resolution images. However, in the current digital watermarking technology, usually small-capacity watermarks are embedded and extracted on low-resolution images, although its performance has achieved good results, its technology is difficult to be transferred to the high-resolution high-capacity watermarking scenario. Therefore, how to break through the large-capacity watermarking technology is a topic worthy of research.

In deep learning-based watermarking technology \cite{Zhu_2018_ECCV,Tancik_2020_CVPR,10155247,jia2021mbrs,9956019,ma2022towards,fang2023denol,10098654,guo2023practical,wu2026aaai,10705361}, the most representative method is HiDDeN, where Zhu \cite{Zhu_2018_ECCV} et al. first propose a complete end-to-end watermarking framework based on convolutional neural networks. This framework supports an image resolution of 128×128 and a watermark capacity of 30 bits. The framework consists of an encoder, a noise layer, a decoder, and a discriminator. The encoder embeds the watermark into the original image to produce the watermarked image, the noise layer simulates various noise attacks, the decoder extracts the watermark from the noisy image, and the discriminator distinguishes as much as possible between the original image and the watermarked image, thereby improving imperceptibility. Stegastamp is proposed by Tancik \cite{Tancik_2020_CVPR} et al. and presents a watermarking framework based on U-Net \cite{ronneberger2015u}. It supports an image resolution of 400×400 and a watermark capacity of 100 bits. This framework primarily improves robustness by refining the embedding position of the watermark in the encoder and the design of the noise layer. ARWGAN is proposed by Huang \cite{10155247} et al. and introduces an attention-guided robust image watermarking model based on generative adversarial networks \cite{goodfellow2020generative,yu2020attention}. It supports an image resolution of 128×128 and a watermark capacity of 30 bits. The framework mainly achieves multi-layer feature extraction and watermark embedding through a feature fusion module and an attention module, effectively enhancing the robustness of the algorithm. MBRS is proposed by Jia \cite{jia2021mbrs} et al. and addresses issues such as the inability to backpropagate gradients. It supports an image resolution of 256×256 and a watermark capacity of 256 bits. This method mainly adopts a mini-batch strategy, where in each small training batch, one of three options—undistorted identity, real JPEG, or simulated JPEG, is randomly selected as the noise layer. The characteristics of the optimizer ensure gradient backpropagation in the majority of cases. DeEND is proposed by Fang \cite{9956019} et al. and presents a decoder-driven watermarking framework. It supports an image resolution of 128×128 and a watermark capacity of 64 bits. In this framework, before the original image and watermark are input into the encoder, they are first passed through the decoder to obtain an intermediate image, which is then fed into the encoder. This approach further improves both the visual quality and robustness of the generated watermarked image. CIN is proposed by Ma \cite{ma2022towards} et al. and introduces a watermarking framework that combines reversible and irreversible mechanisms. It supports an image resolution of 128×128 and a watermark capacity of 30 bits. The framework enhances algorithm performance by designing diffusion and extraction modules, reversible modules, fusion and segmentation modules, irreversible-based attention modules, and specific noise selection modules. DeNoL is proposed by Fang \cite{fang2023denol} et al. and presents a cross-channel robust watermarking network. It supports an image resolution of 128×128 and a watermark capacity of 30 bits. Through a decoupled noise layer with cross-channel simulation, this network is suitable for few-shot learning scenarios. Adaptor is proposed by Wang \cite{10098654} et al. and introduces a separately trained adaptive watermarking method. It supports an image resolution of 128×128 and a watermark capacity of 64 bits. This method primarily adds an adaptor that automatically selects intensity factors to control the embedding strength of the watermark, further improving the visual quality and robustness of the watermarked image. ARIW is proposed by Wu \cite{wu2026aaai} et al. and presents an iterative watermarking framework. It supports an image resolution of 400×400 and a watermark capacity of 100 bits. Through a parallel optimization approach, this framework improves visual quality and robustness to a certain extent. PEE is proposed by Liu \cite{liu2025post} et al. and introduces a screen-shooting robust watermarking method based on feature enhancement and hybrid distortion simulation. It supports an image resolution of 128×128 and a watermark capacity of 64 bits. This framework adopts a two-stage encoder structure. It embeds the watermark in imperceptible regions of the image using a pre-encoder to ensure imperceptibility, then separates and processes frequency components through an enhancement network. It combines convolution and self-attention mechanisms \cite{huang2023understanding} to enhance the robustness of low-frequency components. WOFA is proposed by Liu \cite{liu2025watermarking} et al. and presents a digital watermarking method designed to resist local content theft. It supports an image resolution of 200×200 and a watermark capacity of 30 bits. This method employs a multi-stage network structure and a two-stage training strategy to optimize watermarking performance. STDCN is proposed by Ma \cite{ma2024geometric} et al. and introduces a watermarking model architecture. It supports a watermarked image resolution of 224×224 and a watermark capacity of 196 bits. This architecture combines Swin-Transformer \cite{liu2021swin} and deformable convolutional networks \cite{dai2017deformable}, effectively improving the network's robustness against various geometric distortions. TEEN is a robust watermarking framework with synchronization characteristics proposed by Wang \cite{10705361} et al., which supports a image resolution of 512×512 and a watermark capacity of 40 bits. The template enhancement extraction network proposed by this framework can effectively extract the deformation template from deformed watermarked images and predict the attack factors based on the deformation template, thereby restoring the watermark state and achieving watermark synchronization \cite{li2026self} and watermark extraction.


In summary, the current watermarking methods mainly focus on the research of low-capacity watermarks (e.g., lengths of 30, 100, 256 bits, etc.) in low-resolution images (e.g., 256×256, 400×400, 512×512, etc.). The embedding of low-capacity watermarks results in more redundant information in the watermarked images, which naturally endows these methods with exceptional performance in terms of visual quality and robustness against noise attacks. However, due to their limited embedding capacity, to overcome the constraints of existing robust watermarking techniques, this paper will focus on the research of high-capacity robust watermarking techniques for high-resolution images, while ensuring that the visual quality and robustness of the watermarked image meet the requirements of the objective scenarios. The main contributions are as follows:

\begin{itemize}
	\item \sloppy A high-capacity robust watermarking method for high-resolution images is proposed. Through a block-wise strategy, it effectively embeds high-capacity watermarks into local regions, enabling the network to train normally even under low-resource conditions.
	\item The loss function adopts a weighted combination of global loss and local loss. By optimizing the visual quality and robustness of local block regions, the imperceptibility and robustness of the overall image are further enhanced.
	\item Extensive experiments validate the effectiveness of the proposed method, demonstrating its outstanding performance in embedding high-capacity watermarks, generating high visual quality, and resisting various noise attacks.
\end{itemize}
\begin{figure*}[!t]
	\centering
	\includegraphics[width=6.in]{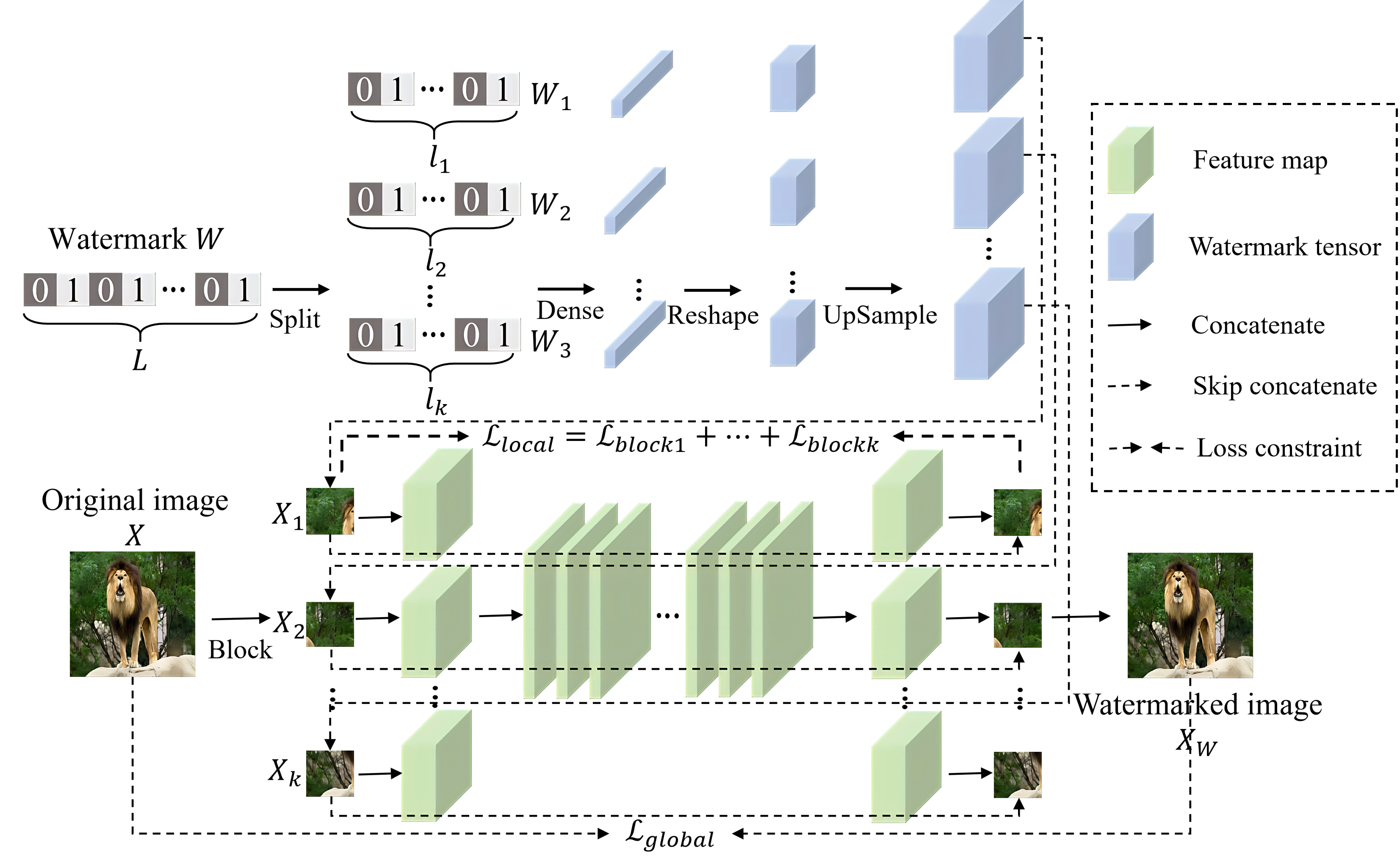}
	\caption{The encoder architecture of the proposed method. 
	}
	\label{fig1}
\end{figure*}
\section{Motivation}
\subsection{Analysis of High-Capacity Watermarks}
Typically, previous watermarking schemes primarily focus on fixed-length, low-capacity watermarks, such as watermark lengths of 30, 100, or 256 bits. After a low-capacity watermark is embedded into a carrier image, the relatively small amount of information it carries allows more redundant watermark information to be embedded into the carrier image, thereby improving the accuracy of watermark extraction. In contrast, high-capacity watermarks carry a larger amount of information, resulting in less redundancy when embedded into the carrier image, which has a certain impact on the accuracy of watermark extraction. However, in real world scenarios, there is sometimes a need to embed a large amount of watermark information, such as text information consisting of hundreds, thousands, or tens of thousands of characters. In such cases, the robustness, imperceptibility, and embedding capacity performance are critically tested. To investigate high-capacity watermarking technology, this paper attempts to decompose high-capacity watermarks, achieving the goal of high-capacity watermark embedding through the embedding of local low-capacity watermarks. Meanwhile, to enhance the adaptability of the watermarking scheme, the designed scheme will support the embedding and extraction of watermarks of arbitrary length, as long as the embedded watermark length is less than 4KB.
\subsection{Analysis of High-Resolution Images}
High-resolution images often lead to a surge in network learning parameters, which typically makes it difficult to train and learn network parameters under resource-constrained conditions. Therefore, training high-capacity watermarking techniques with strong robustness for high-resolution images in low-resource scenarios is highly challenging. To investigate high-capacity robust watermarking technology for high-resolution images, this paper attempts to partition high-resolution images into local blocks. By dividing the image into blocks, the dimensionality of the image is reduced, thereby decreasing the number of parameters in convolutional layers or fully connected layers within the network. Meanwhile, to correspond with high-capacity watermarks, each local region after partitioning will correspond to a low-capacity watermark, ensuring that each embedded watermark and its corresponding region are independent of one another. This approach significantly reduces resource utilization, enabling the network to train and learn even with limited GPU resources. However, blocking also leads to a reduction in the correlation between local regions. In other words, the correlation between local watermarks is poor, and the accuracy of extracting watermarks is limited to a certain extent.
\begin{figure*}[!t]
	\centering
	\includegraphics[width=6.in]{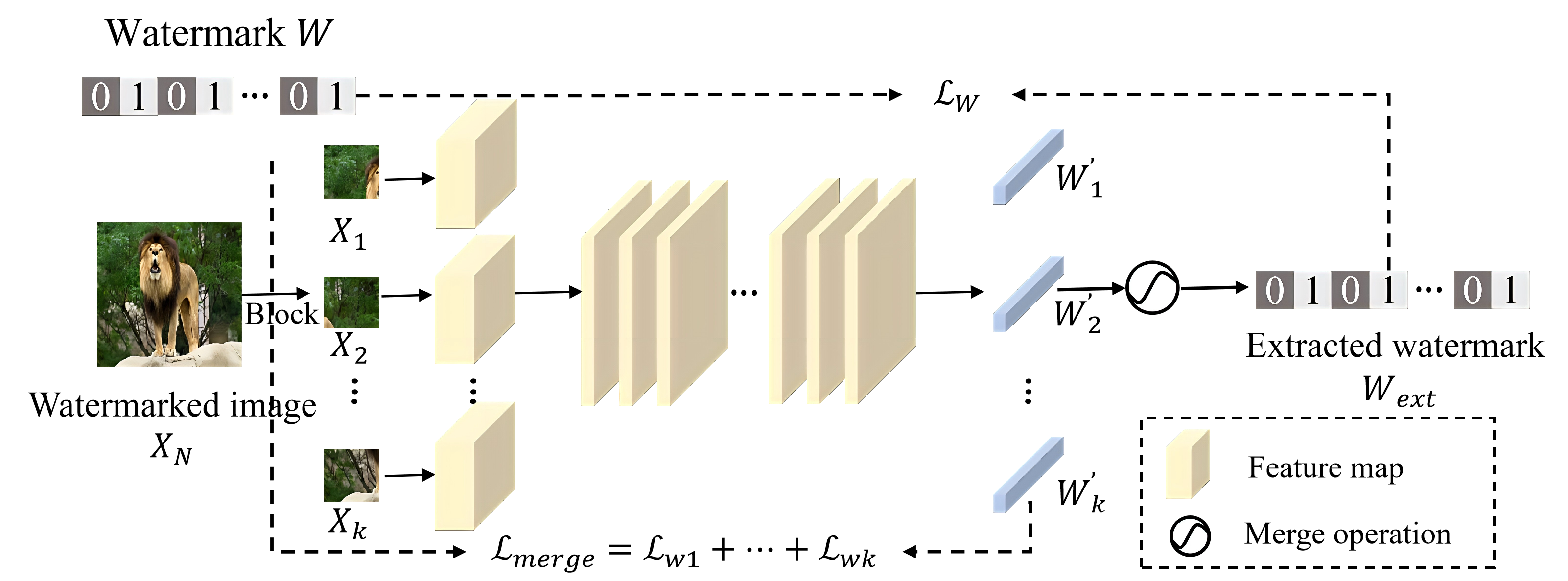}
	\caption{The decoder architecture of the proposed method. 
	}
	\label{fig2}
\end{figure*}
\section{Method}
\subsection{Overview}
This paper proposes a high-capacity robust watermarking method for high-resolution images. This method adapts by partitioning the image into local blocks and matching them with the watermark, thereby transforming the problem of embedding high-capacity watermarks in high-resolution images into multiple problems of embedding low-capacity watermarks in low-resolution images. At the same time, this method significantly reduces the number of network learning parameters, enabling network training even under low-resource conditions. The main modules of this method include two components, namely the encoder and the decoder. However, to enhance the ability to resist noise attacks, this method also inserts a noise layer between the encoder and decoder to simulate noise attacks. The overall scheme adopts a joint training approach to optimize and learn the loss function. The encoder and decoder structures of the proposed method are shown in Fig. \ref{fig1} and Fig. \ref{fig2}, respectively, and the modules within these structures will be described in detail below.

\subsection{Encoder}
\label{sec:subsubsection}
The main task of the encoder $E$ is to embed the watermark $W\in\{0,1\}^L$ into the original image $X\in\mathbb{R}^{m\times n\times c}$ in an invisible form and generate the watermarked image $X_W\in\mathbb{R}^{m\times n\times c}$. To simultaneously address the requirements of high-capacity watermark, imperceptibility, and robustness, this paper processes the watermark and the image using segmentation and blocking methods, respectively, enabling network adaptation. Specifically, the original watermark $W$ is first segmented into $k$ mutually independent watermark segments $W_i\in\{0,1\}^{l_i}, i=1,2,...,k$. Each segment then undergoes a linear transformation, followed by a reshape operation to transform its dimensions, and finally an upsampling operation to obtain the watermark tensor to be embedded. The process is shown in Fig. \ref{fig1}. For the corresponding high-resolution image, the image $X$ is first divided into blocks $X_i\in\mathbb{R}^{m_i\times n_i\times c_i}, i=1,2,...,k$, ensuring that the number of image blocks matches the total number of watermark tensors. Each watermark tensor is then associated with its corresponding image block and fed into the network for learning. In this process, each watermark tensor and image block pair undergoes parameter learning using the same network structure. The network structure employed consists of three convolutional layers and is designed to be reversible and symmetric with respect to the decoder $D$ structure. The watermark embedding process by encoder $E$ can be described as follows:
\begin{equation}\label{encoder_1}
	\begin{split}
		X_W=E(X,W)=merge(\sigma(\mathcal{K}(X_1\textcircled{c}W_1)),...,\sigma(\mathcal{K}(X_k\textcircled{c}W_k)))
	\end{split}
\end{equation}
where $\sigma$ is the activation function $Relu = max(0, x)$, $\mathcal{K}$ is the convolution operator, $k$ is the number of image blocks, $\textcircled{c}$ is the concatenation operation and $merge$ is the aggregation operation. It is worth noting that $\sigma(\mathcal{K}(X_i\textcircled{c}W_i))$ represents the entire watermark embedding process for the local image block within the encoder, and the watermark tensor is coupled at each convolutional layer. Naturally, in the decoder, watermark features can be decoupled from each convolutional layer and subsequently fused to improve watermark extraction accuracy.

On the other hand, to ensure the imperceptibility, this method imposes both local and global visual quality constraints on the original image $X$ and the watermarked image $X_W$. To address the robustness requirement, similar to existing methods, this approach also employs a noise layer $N$ to simulate noise attacks. The noise layer $N$ is inserted between the encoder $E$ and the decoder $D$, taking the watermarked image $X_W$ as input and producing the noisy watermarked image $X_N$ as output. In the design of the noise layer, this method adopts a cascading approach to simulate multiple types of noise attacks sequentially. For example, the first noise attack simulates JPEG compression on the watermarked image $X_W$, followed by a second attack simulating Gaussian noise, and so on, until the sequence of simulated noise attacks is complete. This manner of simulating noise attacks effectively corresponds to emulating a new combined noise attack. The process of simulating noise attacks using a noise layer $N$ can be represented as follows:
\begin{equation}\label{encoder_1}
	\begin{split}
		X_N=N(X_W)=N_s(N_{s-1}(...N_1(X_w)...))
	\end{split}
\end{equation}
where $N_i$ represents the attack function for the i-th type of noise.

\begin{figure}[!h]
	\centering
	\includegraphics[width=2.in]{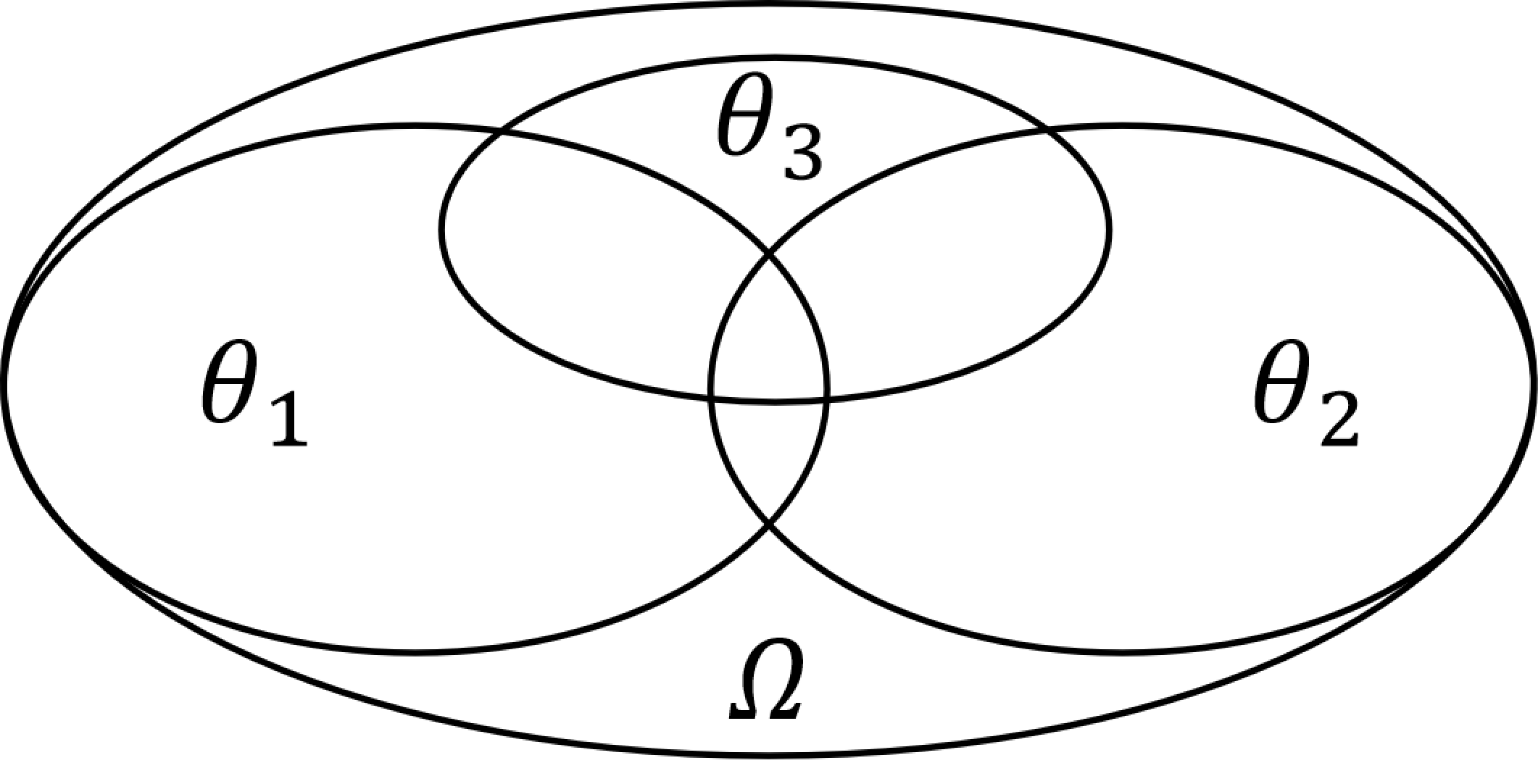}
	\caption{A schematic diagram illustrating the robustness boundaries of pixel value modifications caused by different noise attacks when $s=3$. $\Omega$ represents the optimal robustness boundary that each watermarking scheme ultimately needs to optimize.}
	\label{fig3}
\end{figure}

In order to facilitate the description of the robustness caused by each noise, let the upper bound of the robustness boundary for pixel value modifications caused by each individual noise attack be denoted as the set $\theta_i,i=1,2,...,s$, $s$ represents the number of noise types, and the ideal lower bound of the robustness boundary be denoted as the set $\Omega$. Consequently, the robustness boundaries corresponding to these noise attacks can be illustrated as shown in Fig. \ref{fig3}. Fig. \ref{fig3} is a schematic diagram illustrating the robustness boundaries of pixel value modifications caused by different noise attacks when $s=3$. $\theta_1$ represents the robustness boundary corresponding to the first noise attack, and $\theta_2$ represents the robustness boundary corresponding to the second noise attack. There is an intersection in the pixel value modifications caused by these two types of noise, so their robustness boundaries also exist for intersection $\theta_1\cap\theta_2$. The robustness boundaries corresponding to other noise attacks can be analyzed similarly. $\Omega$ is the optimal lower bound when robustness is satisfied for all noise attacks:
\begin{equation}\label{encoder_1}
	\begin{split}
		\Omega=\theta_1\cup\theta_2\cup...\cup\theta_s
	\end{split}
\end{equation}

The smaller $\Omega$ set is, the better the imperceptibility. To balance visual quality and robustness, the network trained for the watermarking scheme should strive to make $\Omega$ as small as possible. This implies a reduction in the amount of modification to image pixel values, while ensuring that the robustness covers all types of noise attacks.

\subsection{Decoder}
The main task of the decoder $D$ is to extract the watermark from the noisy image $X_N$. Since the watermarked image $X_W$ has been subjected to noise attacks, its watermark may have been modified. To extract the watermark from the noisy image as accurately as possible, this method employs a reversible network that is symmetric to the encoder structure, as shown in Fig. \ref{fig2}. First, the noisy image $X_N$ is divided into blocks $X_i\in\mathbb{R}^{m_i\times n_i\times c_i}, i=1,2,...,k$. Each image block then undergoes parameter learning using the same network structure, and the watermark is decoupled and separated from each convolutional feature map. The $k$ image blocks yield $k$ watermark segments, which are finally integrated to obtain the final extracted watermark $W_{ext}$. The watermark extraction process by the decoder can be represented as follows:
\begin{equation}\label{decoder_1}
	\begin{split}
		W_{ext}=D(X_N)&=merge(W^{\prime}_1,...,W^{\prime}_k)\\
		&=merge(\sigma(\mathcal{K}^{-1}(X_1)),...,\sigma(\mathcal{K}^{-1}(X_k)))
	\end{split}
\end{equation}
where $\mathcal{K}^{-1}$ denotes the deconvolution operator, and $\sigma(\mathcal{K}^{-1}(X_i))$ represents the entire watermark extraction process for the local image block within the decoder, where watermark features are decoupled at each convolutional layer. These watermark features, after being fused, are passed through a fully connected layer with a sigmoid activation function, and its output corresponds to the length $l_i$ associated with each watermark $W_i$.

To improve the accuracy of watermark extraction, the decoder primarily uses watermark cross-entropy \cite{yin2023anti} to measure its performance. In constraining the watermark loss, this method adopts both local and global watermark loss constraints. This enables the network to focus on high-capacity embedding and visual quality, while also balancing the network's robustness against noise attacks.

\begin{table*}[!t]
	\centering
	\caption{The PSNR and SSIM of three test sets at different embedding strengths.}
	\label{tab1}
	\begin{tabular}{lrrrrrrrrrrr}
		\toprule
		Dataset & Metric & 0.2 & 0.4 & 0.6 & 0.8 & 1.0 & 1.2 & 1.4 & 1.6 & 1.8 & 2.0 \\
		\midrule
		Mirflickr & PSNR$\uparrow$ &49.59 &44.61 &41.40 &39.04 &37.19 &35.67 &34.38 &33.25 &32.26 &31.37\\
		& SSIM$\uparrow$ &0.9947 &0.9834 &0.9669 &0.9472 &0.9258 &0.9040 &0.8823 &0.8611 &0.8408 &0.8215\\
		\midrule
		COCO & PSNR$\uparrow$ &49.25 &44.20 &40.96 &38.60 &36.74 &35.21 &33.90 &32.77 &31.77 &30.88\\
		& SSIM$\uparrow$ &0.9950 &0.9840 &0.9681 &0.9485 &0.9270 &0.9046 &0.8822 &0.8603 &0.8393 &0.8193\\
		\midrule
		BOSSBase & PSNR$\uparrow$ &49.92 &45.09 &41.91 &39.55 &37.69 &36.15 &34.85 &33.73 &32.74 &31.85\\
		& SSIM$\uparrow$ &0.9948 &0.9838 &0.9677 &0.9480 &0.9264 &0.9042 &0.8822 &0.8610 &0.8408 &0.8217\\
		\bottomrule
	\end{tabular}
\end{table*}
\begin{figure*}[!t]
	\centering
	\includegraphics[width=\linewidth]{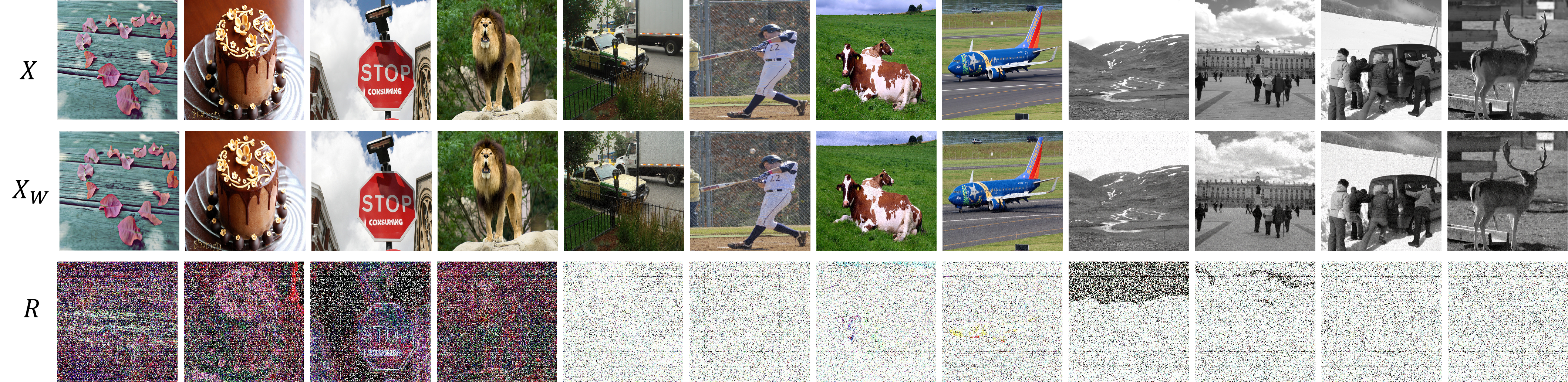}
	\caption{Visualization of the partial watermarked image generated by the proposed method. The first row represents the original image $X$, the second row represents the corresponding watermarked image $X_W$, and the third row is the corresponding residual image $R =abs(X_W-X)$. The first four columns of images are from the Mirflickr database, the middle four columns are from the COCO database, and the last four columns are from the BOSSBase database.}
	\label{fig4}
\end{figure*}
\subsection{Loss Function}
Designing the loss function is crucial for the success of network model optimization. For the encoder $E$, its main objective is to generate high-quality watermarked images; therefore, this part can be achieved by constraining the changes in the embedding regions before and after watermark embedding. For the decoder $D$, its primary purpose is to accurately extract watermark with a certain degree of generalization. Thus, this part is mainly realized by constraining the relationship between the extracted watermark and the original watermark. Since the embedding regions correspond to low-resolution images, for high-resolution images, we need to constrain each individual embedding region rather than solely adopting a global constraint approach. Meanwhile, for each embedding region, the loss for watermark extraction is designed separately, rather than using only the loss after fusing features from all regions. This design aligns well with the transformation of high-resolution regions into low-resolution regions. 

Therefore, the final loss function in this paper consists of four components: the global image loss $\mathcal{L}_{global}$, the local image loss $\mathcal{L}_{local}$, the global watermark loss $\mathcal{L}_{W}$, and the local watermark loss $\mathcal{L}_{merge}$. $\mathcal{L}_{global}$ and $\mathcal{L}_{local}$ are constrained by calculating the mean squared error (MSE) \cite{wang2004image}. Let $X_E$ denote the region to be embedded, and $X^\prime_E$ denote the region after watermark embedding. Then, the calculation formulas for 
$\mathcal{L}_{global}$ and $\mathcal{L}_{local}$ are as follows:
\begin{equation}\label{loss_1}
	\begin{split}
		\mathcal{L}_{global}=MSE(X,X_W)=\frac{1}{m\times n\times c}\sum_{i=1}^{m\times n\times c}(x_i-x^{\prime}_i)^2
	\end{split}
\end{equation}
\begin{equation}\label{loss_1}
	\begin{split}
		\mathcal{L}_{local}=\sum^k_{i=1}\mathcal{L}_i=\sum^k_{i=1}MSE(X_{E_i},X^{\prime}_{E_i})
	\end{split}
\end{equation}
Where $m$, $n$and $c$ are the width, height, and number of channels of the image $X$, respectively. $x_i$and $x^{\prime}_i$ are the pixel values at the same coordinates in the image $X$ and $X_W$, respectively. $k$ is the number of embedding regions.

$\mathcal{L}_{W}$ and $\mathcal{L}_{merge}$ are constrained by calculating the cross-entropy \cite{yin2023anti}. Let $W_i$ denote the watermark of the i-th segment, and $W^{\prime}_i$ be the watermark extracted from the i-th image block $X_i$. Then, the calculation formulas fo $\mathcal{L}_{W}$and $\mathcal{L}_{merge}$are as follows:
\begin{equation}\label{loss_1}
	\begin{split}
		\mathcal{L}_{W}=BCH(W,W_{ext})=\frac{1}{L}\sum_{i=1}^{L}-w_i\log w^{\prime}_i-(1-w_i)\log (1-w^{\prime}_i)
	\end{split}
\end{equation}
\begin{equation}\label{loss_1}
	\begin{split}
		\mathcal{L}_{merge}=\sum^k_{i=1}\mathcal{L}_i=\sum^k_{i=1}BCE(W_i,W^{\prime}_i)
	\end{split}
\end{equation}
Therefore, the final loss function of the network is:
\begin{equation}\label{loss_5}
	\mathcal{L}_{Total}=\lambda_1\mathcal{L}_{global}+\lambda_2\mathcal{L}_{local}+\lambda_3\mathcal{L}_{W}+\lambda_4\mathcal{L}_{merge}
\end{equation}
where $\lambda_1$, $\lambda_2$, $\lambda_3$ and $\lambda_4$ are the corresponding loss weights, whose initial values are $(1.0,1.0,1.0,1.0)$. These weights can be adjusted during training to balance the contributions of each loss component.

\begin{table}[!t]
	\centering
	\caption{Robustness under different JPEG compression quality factors (\%).}
	\label{tab2}
	\begin{tabular}{lrrrrrr}
		\toprule
		Dataset & 50 & 60 & 70 & 80 &90 &100\\
		\midrule
		Mirflickr & 98.51&99.34 &99.79 &99.94 &99.98&99.99\\
		\midrule
		COCO &98.60 &99.39 &99.80 &99.94 &99.98&99.99\\
		\midrule
		BOSSBase &98.72 &99.48 &99.85 &99.97 &99.99&100.0\\
		\bottomrule
	\end{tabular}
\end{table}
\begin{table}[!t]
	\centering
	\caption{Robustness under different variances of Gaussian noise (\%).}
	\label{tab3}
	\begin{tabular}{lrrrr}
		\toprule
		Dataset & 0.02 & 0.01 & 0.001 & 0.0001\\
		\midrule
		Mirflickr & 99.66&99.97 &99.99 &99.99 \\
		\midrule
		COCO &99.81 &99.98 &99.99 &99.99 \\
		\midrule
		BOSSBase &99.93 &99.99 &100.0 &100.0 \\
		\bottomrule
	\end{tabular}
\end{table}
\begin{table}[!t]
	\centering
	\caption{Robustness under other noise attacks (\%).}
	\label{tab4}
	\resizebox{\linewidth}{!}{\begin{tabular}{lrrrrrr}
			\toprule
			Dataset & Identity & \makecell{Dropout} & \makecell{Cropout} & \makecell{Crop} &\makecell{Gaussian\\filtering} &\makecell{Scaling}\\
			\midrule
			Mirflickr & 100.0&100.0 &99.99 &97.36 &99.99&97.06\\
			\midrule
			COCO &100.0 &99.99 &99.99 &96.83 &99.99&96.75\\
			\midrule
			BOSSBase &100.0 &100.0 &100.0 &97.19 &100.0&97.68\\
			\bottomrule
	\end{tabular}}
\end{table}

\begin{table*}[!t]
	\centering
	\caption{The embedding capacity, visual quality and robustness performance of different watermarking methods (\%).}
	\label{tab5}
	\resizebox{\linewidth}{!}{\begin{tabular}{c c c c c c c c c c c c c c c}
			\toprule
			Dataset &Method &$m\times n$ &$L$&ER$\uparrow$&PSNR$\uparrow$&SSIM$\uparrow$& Identity & \makecell{JPEG\\(QF=50)} & \makecell{Gaussian\\noise\\($\sigma$=0.02)} & \makecell{Gaussian\\filter\\($k$=7)} & \makecell{Crop\\($p$=0.03)} & \makecell{Cropout\\($p$=0.9)} & \makecell{Dropout\\($p$=0.9)} & \makecell{Scaling\\($p$=0.5)}\\
			\midrule
			\multirow{8}{*}{Mirflickr}&HiDDeN\cite{Zhu_2018_ECCV}&$128\times128$&30&0.0018&32.56&0.9373&93.75&91.83&97.76&88.03&75.37&91.53&97.50&78.23\\
			&StegaStamp\cite{Tancik_2020_CVPR}&$400\times400$&100&0.0006&29.30&0.8900&99.93&99.89&99.84&99.92&99.61&99.32&99.85&99.84\\
			&MBRS\cite{jia2021mbrs}&$256\times256$&256&0.0039&41.99&0.9880&100.0&99.36&81.12&55.47&49.78&94.22&100.0&49.93\\
			&CIN\cite{ma2022towards}&$128\times128$&30&0.0018&41.44&0.9800&100.0&99.67&100.0&100.0&99.93&100.0&98.30&98.93\\
			&ARWGAN\cite{10155247}&$128\times128$&30&0.0018&22.65&0.8770&99.16&90.27&94.50&99.13&95.80&98.87&99.23&95.83\\
			&STDCN\cite{ma2024geometric}&$224\times224$&196&0.0034&35.97&0.9920&100.0&99.93(40)&98.09(0.04)&100.0&96.75(0.2)&96.87(0.5)&95.44(0.7)&99.38(0.4)\\
			&ARIW\cite{wu2026aaai}&$400\times400$&100&0.0006&42.08&0.9870&100.0&99.66&99.96&99.99&99.71&99.98&99.99&99.94\\
			&\cellcolor{gray!20}\textbf{\underline{Ours}}&\cellcolor{gray!20}$1024\times1024$&\cellcolor{gray!20}32768&\cellcolor{gray!20}0.0313&\cellcolor{gray!20}37.19&\cellcolor{gray!20}0.9258&\cellcolor{gray!20}100.0 &\cellcolor{gray!20}98.51 &\cellcolor{gray!20}99.66 &\cellcolor{gray!20}99.99 &\cellcolor{gray!20}97.36 &\cellcolor{gray!20}99.99 &\cellcolor{gray!20}100.0 &\cellcolor{gray!20}97.06\\
			\midrule
			\multirow{8}{*}{COCO}&HiDDeN\cite{Zhu_2018_ECCV}&$128\times128$&30&0.0018&32.89&0.9470&98.33&92.73&97.93&89.93&74.22&92.23&98.23&79.13\\
			&StegaStamp\cite{Tancik_2020_CVPR}&$400\times400$&100&0.0006&30.07&0.9100&99.88&99.82&99.88&99.88&99.62&99.18&99.79&99.90\\
			&MBRS\cite{jia2021mbrs}&$256\times256$&256&0.0039&42.01&0.9880&100.0&99.62&79.57&100.0&49.61&93.96&100.0&50.36\\
			&CIN\cite{ma2022towards}&$128\times128$&30&0.0018&41.91&0.9830&100.0&99.77&100.0&100.0&99.97&100.0&97.40&99.27\\
			&ARWGAN\cite{10155247}&$128\times128$&30&0.0018&31.74&0.9590&99.20&94.86&96.13&99.20&96.23&99.23&99.06&96.30\\
			&STDCN\cite{ma2024geometric}&$224\times224$&196&0.0034&35.97&0.9920&100.0&99.93(40)&98.09(0.04)&100.0&96.75(0.2)&96.87(0.5)&95.44(0.7)&99.38(0.4)\\
			&ARIW\cite{wu2026aaai}&$400\times400$&100&0.0006&41.76&0.9880&100.0&99.54&99.98&99.98&99.88&99.98&99.99&99.94\\
			&\cellcolor{gray!20}\textbf{\underline{Ours}}&\cellcolor{gray!20}$1024\times1024$&\cellcolor{gray!20}32768&\cellcolor{gray!20}0.0313&\cellcolor{gray!20}36.74&\cellcolor{gray!20}0.9270&\cellcolor{gray!20}100.0 &\cellcolor{gray!20}98.60 &\cellcolor{gray!20}99.81 &\cellcolor{gray!20}99.99 &\cellcolor{gray!20}96.83 &\cellcolor{gray!20}99.99 &\cellcolor{gray!20}99.99 &\cellcolor{gray!20}96.75\\
			\midrule
			\multirow{8}{*}{BOSSBase}&HiDDeN\cite{Zhu_2018_ECCV}&$128\times128$&30&0.0018&35.94&0.9700&97.00&87.27&98.76&88.77&75.70&92.77&98.47&77.53\\
			&StegaStamp\cite{Tancik_2020_CVPR}&$400\times400$&100&0.0006&30.07&0.9100&99.99&99.97&99.92&99.97&99.75&99.60&99.93&99.91\\
			&MBRS\cite{jia2021mbrs}&$256\times256$&256&0.0039&42.98&0.9880&100.0&99.87&78.99&99.30&49.24&94.42&100.0&50.18\\
			&CIN\cite{ma2022towards}&$128\times128$&30&0.0018&43.22&0.9840&100.0&99.60&100.0&100.0&100.0&100.0&96.40&99.30\\
			&ARWGAN\cite{10155247}&$128\times128$&30&0.0018&31.74&0.9590&95.87&85.93&93.09&99.30&97.20&99.17&98.993&97.70\\
			&STDCN\cite{ma2024geometric}&$224\times224$&196&0.0034&35.97&0.9920&100.0&99.93(40)&98.09(0.04)&100.0&96.75(0.2)&96.87(0.5)&95.44(0.7)&99.38(0.4)\\
			&ARIW\cite{wu2026aaai}&$400\times400$&100&0.0006&42.02&0.9870&100.0&99.80&99.99&100.0&99.97&100.0&100.0&100.0\\
			&\cellcolor{gray!20}\textbf{\underline{Ours}}&\cellcolor{gray!20}$1024\times1024$&\cellcolor{gray!20}32768&\cellcolor{gray!20}0.0313&\cellcolor{gray!20}37.69&\cellcolor{gray!20}0.9264&\cellcolor{gray!20}100.0 &\cellcolor{gray!20}98.72 &\cellcolor{gray!20}99.93 &\cellcolor{gray!20}100.0 &\cellcolor{gray!20}97.19 &\cellcolor{gray!20}100.0 &\cellcolor{gray!20}100.0 &\cellcolor{gray!20}97.68\\
			\bottomrule
	\end{tabular}}
\end{table*}
\section{Experiment}
\subsection{Experimental Setup}
\subsubsection{Dataset}In this paper, the training dataset consists of 2,000 randomly selected color images from the Mirflickr \cite{huiskes2008mir} database. The testing set includes three datasets: 100 randomly selected color images from the Mirflickr database, 100 randomly selected grayscale images from the BOSSBase \cite{bas2011break} database, and 100 randomly selected color images from the COCO \cite{lin2014microsoft} database. Before use, all these images are uniformly rescaled to a resolution of 1024×1024 to maintain consistency with the input and output requirements of this method.
\subsubsection{Evaluation Metric}This paper involves the evaluation of three performance metrics, including embedding capacity, visual quality, and robustness. The embedding capacity is measured using the embedding rate (ER) \cite{liu2025image}, which quantifies the watermark bit capacity embedded per pixel (bpp). A higher ER value indicates that more watermark information is carried per pixel. Visual quality is assessed using two metrics: average peak signal-to-noise ratio (PSNR) and structural similarity (SSIM) \cite{wang2004image}. Robustness is measured by the watermark extraction accuracy, which includes the accuracy of watermark extraction under various types of noise attacks.
\subsubsection{Implementation Detail}The experiments are conducted on a platform running the Windows 10 operating system, equipped with an Intel(R) Xeon(R) Gold 6161 CPU @ 2.20GHz and 128 GB of memory. The programming development environment is Python 3.6. Additionally, the network uses Adam as the optimizer, with a learning rate of 0.001, a batch size of 1, and 140,000 training iterations. The convolution kernel size is 3×3 with a stride of 1. Furthermore, the maximum embedded watermark capacity is fixed at 32,768 bits (4KB), and the embedding strength $\alpha$ is set to 1.0.

\subsection{Visual Quality}
Table \ref{tab1} presents the PSNR and SSIM performance of the proposed method under different embedding strengths. Specifically, when the embedding strength $\alpha=1.0$, the PSNR exceeds 36 dB across all three datasets, and the SSIM exceeds 0.92. It can be observed that in scenarios where greater attention is paid to the visual quality of the watermarked image, the requirement can be met by reducing the embedding strength. For example, when $\alpha=0.6$, the PSNR exceeds 40 dB across all three datasets, and the SSIM exceeds 0.96. Of course, reducing the embedding strength will, to some extent, lead to a decrease in watermark extraction accuracy. Therefore, in practice, an appropriate watermarking scheme should be selected based on specific requirements. Fig. \ref{fig4} presents visualizations of some watermarked images generated by the proposed method. It can be observed that without magnifying the images, the difference between the original images and the watermarked images is imperceptible to the human eye, demonstrating the good imperceptibility of the proposed method. However, when the images are magnified, traces of modification become noticeable in some images, particularly along the edges of image blocks, where the alterations are relatively apparent. Analyzing the reasons, we believe this is related to the experimental setup. In the experimental configuration, we did not use native images with a resolution of 1024×1024; instead, we obtained 1024×1024 images by scaling up lower-resolution images. Across the three databases, the original image resolutions are generally lower than 1024×1024. Consequently, when images are scaled up, certain pixels are directly replicated, resulting in large areas with identical pixel values. This makes these regions relatively smooth and unsuitable for watermark embedding. Therefore, there is still room for improvement in this method, which also represents a direction for future work.

\subsection{Robustness}
Regarding robustness, this paper demonstrates the watermark extraction accuracy under simulations of multiple noise attacks. The types of noise attacks include JPEG compression with quality factors (QF) of 50, 60, 70, 80, 90, and 100; Gaussian noise with variances of 0.0001, 0.001, 0.01, and 0.02; additionally, other noise attacks tested include Identity, Dropout (0.9), Cropout (0.9), Crop (0.03), Gaussian filtering (k=7, var=0.1), and Scaling (0.5), where the numbers in parentheses indicate the corresponding attack intensities. Tables \ref{tab2}, \ref{tab3} and \ref{tab4} show the accuracy rates of watermark extraction for this method under different noise attacks (with $\alpha=1.0$). It can be observed that with an embedding strength of alpha = 1.0, the model achieves an accuracy exceeding 98.51\% under JPEG compression attacks, over 99.66\% under Gaussian noise attacks, more than 99.99\% under Identity, Dropout, Cropout, and Gaussian filtering attacks, and over 96.75\% under Crop and Scaling attacks. The robustness of the model against various noise attacks demonstrates that the proposed method is effective and feasible for high-capacity watermark embedding in high-resolution images.

\begin{table*}[!t]
	\centering
	\caption{Visual quality and robustness of local loss terms $\mathcal{L}_{local}$ and $\mathcal{L}_{merge}$ with or without (\%).}
	\label{tab7}
	\begin{tabular}{ccccccccccccc}
		\toprule
		\multirow{2}{*}{$\mathcal{L}_{global}$} & \multirow{2}{*}{$\mathcal{L}_{local}$} & \multirow{2}{*}{$\mathcal{L}_{W}$} & \multirow{2}{*}{$\mathcal{L}_{merge}$} & \multicolumn{3}{c}{Mirflickr} &\multicolumn{3}{c}{COCO} &\multicolumn{3}{c}{BOSSBase}\\
		\cmidrule(lr){5-7}\cmidrule(lr){8-10}\cmidrule(lr){11-13}
		&&&& PSNR&SSIM &JPEG& PSNR&SSIM & JPEG& PSNR&SSIM &JPEG\\
		\midrule
		$\checkmark$ & $\times$&$\checkmark$ &$\times$ &37.94 &0.9324&81.15&37.60&0.9352&81.92&38.10&0.9303&82.16\\
		\midrule
		$\checkmark$ & $\checkmark$&$\checkmark$ &$\times$ &42.82 &0.9783&60.98&41.96&0.9773&61.84&43.67&0.9785&61.43\\
		\midrule
		$\checkmark$ & $\times$&$\checkmark$ &$\checkmark$ &33.86 &0.8721&99.59&33.65&0.8770&99.62&34.37&0.8752&99.69\\
		\midrule
		\rowcolor{gray!20}
		$\checkmark$ & $\checkmark$&$\checkmark$ &$\checkmark$ &37.19 &0.9258&98.51&36.74&0.9270&98.60&37.69&0.9264&98.72\\
		\bottomrule
	\end{tabular}
\end{table*}
\subsection{Comparative Experiment}
To highlight the strengths and weaknesses of the proposed method, this paper compares its performance with classic watermarking schemes. The compared watermarking schemes include HiDDeN \cite{Zhu_2018_ECCV}, StegaStamp \cite{Tancik_2020_CVPR}, MBRS \cite{jia2021mbrs}, CIN \cite{ma2022towards}, ARWGAN \cite{10155247}, STDCN \cite{ma2024geometric}, and ARIW \cite{wu2026aaai}. It is worth noting that, except for the STDCN method, whose results are taken directly from the original paper, all other methods are evaluated using open-source pre-trained models on the three test datasets used in this paper. Furthermore, without loss of generality, the types of noise attacks tested mainly include JPEG (50), Gaussian noise (0.02), Identity, Dropout (0.9), Cropout (0.9), Crop (0.03), Gaussian filtering (k=7, var=0.1), and scaling (0.5).

Table \ref{tab5} presents the visual quality and robustness of different watermarking methods under various noise attacks, along with their corresponding watermark embedding rates. In terms of PSNR and SSIM metrics, the visual quality of the proposed method is neither the best nor the worst, placing it in a moderate position. Regarding robustness against noise attacks, the proposed method achieves the best performance under certain types of noise attacks but falls short under others. However, in terms of watermark embedding capacity, the proposed method achieves the highest embedding rate, reaching 0.0313 bpp.

In practice, watermarking methods that achieve excellent performance with low-capacity embedding in low-resolution images face significant challenges when transitioning to high-capacity embedding in high-resolution images. This is because high-resolution, high-capacity scenarios lead to a surge in model parameters, making training difficult under low-resource conditions. Therefore, the design of the proposed method provides a relatively effective and feasible technique for high-capacity watermarking in high-resolution images.

\begin{table}[!t]
	\centering
	\caption{Robustness against JPEG compression attacks under different watermark lengths (\%).}
	\label{tab6}
	\resizebox{\linewidth}{!}{\begin{tabular}{lrrrrrrr}
			\toprule
			Dataset & 64 & 128 & 512 & 1024 &8192 &16384&32768\\
			\midrule
			Mirflickr & 100.0&100.0 &99.89 &99.84 &99.43&98.57&98.51\\
			\midrule
			COCO &100.0 &100.0 &99.92 &99.87 &99.51&98.70&98.60\\
			\midrule
			BOSSBase &100.0 &100.0 &99.80 &99.81 &99.47&98.78&98.72\\
			\bottomrule
	\end{tabular}}
\end{table}
\subsection{Ablation Experiment}
\subsubsection{Loss Function Term}
In general, the design of loss functions primarily focuses on global image loss and global watermark loss, with very little attention paid to local region losses. This section conducts ablation experiments on local losses, mainly including $\mathcal{L}_{local}$ and $\mathcal{L}_{merge}$. Table \ref{tab7} presents the visual quality of the model on three different test sets and its robustness under JPEG (QF=50) attack scenarios with different local loss term optimizations. It can be observed that without the local terms $\mathcal{L}_{local}$ and $\mathcal{L}_{merge}$, the model generates watermarked images with relatively good visual quality, but its robustness is poor, with the highest watermark extraction accuracy reaching only 82.16\%. Without the local term $\mathcal{L}_{merge}$, the model pays more attention to optimizing image visual quality while neglecting robustness. This is mainly because the addition of the local image loss $\mathcal{L}_{local}$ in the total loss function $\mathcal{L}_{Total}$ increases the weight of visual quality optimization. Without the local term $\mathcal{L}_{local}$, conversely, the model focuses more on optimizing robustness while neglecting visual quality. This is primarily due to the increased weight of the watermark loss in the total loss function $\mathcal{L}_{Total}$. Therefore, when both local losses $\mathcal{L}_{local}$ and $\mathcal{L}_{merge}$ are included, it can be observed that the model achieves the optimal balance between visual quality and robustness optimization.

\subsubsection{Watermark Length}
Under the condition of the maximum embedding capacity limit, this section mainly tests the watermark extraction accuracy corresponding to different watermark lengths. Without loss of generality, this section mainly demonstrates the robustness in the JPEG (QF=50) scenario. Here, the watermark lengths are selected as 64 bits, 128 bits, 512 bits, 1024 bits, 8192 bits (1KB), 16384 bits (2KB), and 32768 bits (4KB). Table \ref{tab6} shows the robustness of different watermark capacities. It can be found that when the embedding watermark capacity decreases, its robustness will gradually improve. Reducing the embedding watermark capacity will lead to an increase in the redundancy of the watermark information in the embedded image, thereby improving the watermark extraction accuracy to a certain extent. The method proposed supports the embedding of watermarks of arbitrary length, with a maximum embedding capacity of 4KB, which can meet the requirements of various practical application scenarios. Moreover, when the demand for embedding capacity is not high, the robustness of this method is significantly enhanced.

\section{Conclusion}
To address the challenges of high-capacity watermarking, generated visual quality, and robustness in high-resolution image scenarios, this paper proposes a high-capacity watermarking technique for high-resolution images. Specifically, this paper partitions high-resolution images and high-capacity watermarks to adapt to the training and optimization of the network structure, thereby reducing network parameters and resource utilization. The adopted symmetric reversible network ensures rapid synchronization and extraction of the watermark. Finally, visual quality and robustness are optimized through the constraints of global and local losses. Extensive experiments demonstrate that the proposed method achieves satisfactory performance in terms of large capacity, visual quality, and robustness. Meanwhile, ablation studies validate the effectiveness and feasibility of the network structure design. In summary, this paper provides an effective and feasible solution for transitioning from low-capacity watermarking techniques in low-resolution images to high-capacity watermarking techniques in high-resolution images.

\bibliographystyle{ACM-Reference-Format}
\bibliography{sample-base}

\appendix
\end{document}